# The use of conflicts in searching Bayesian networks

David Poole*
Department of Computer Science,
University of British Columbia,
Vancouver, B.C., Canada V6T 1Z2
poole@cs.ubc.ca

## Abstract

This paper discusses how conflicts (as used by the consistency-based diagnosis community) can be adapted to be used in a search-based algorithm for computing prior and posterior probabilities in discrete Bayesian Networks. This is an "anytime" algorithm, that at any stage can estimate the probabilities and give an error bound. Whereas the most popular Bayesian net algorithms exploit the structure of the network for efficiency, we exploit probability distributions for efficiency; this algorithm is most suited to the case with extreme probabilities. This paper presents a solution to the inefficiencies found in naive algorithms, and shows how the tools of the consistency-based diagnosis community (namely conflicts) can be used effectively to improve the efficiency. Empirical results with networks having tens of thousands of nodes are presented.

## 1  Introduction

There have been two, previously disparate, communities working on model-based diagnosis. The first is in the UAI community, where Bayesian networks have become the representation of choice for modelling. The second is the community built on logic-based notions of diagnosis, and is typified by the use of consistency-based diagnosis [Genesereth, 1984; Reiter, 1987; de Kleer and Williams, 1987; de Kleer *et al.*, 1990].

The basis of consistency based diagnosis is the use of the *conflict* [Reiter, 1987; de Kleer and Williams, 1987; de Kleer *et al.*, 1990]. A conflict is a set of assumptions, the conjunction of which is inconsistent with the observations and the system description. The model-based diagnosis community has recently seen the need for the use of probabilities in reducing the combinatorial explosion in the number of logical possibilities [de Kleer and Williams, 1987; de Kleer, 1991]. This brings their work closer to that of the uncertainty community. The efficiency of these algorithms, and the other issues faced by this community (e.g., the use of abstraction [Genesereth, 1984]) mean that their work cannot be ignored by the uncertainty community.

This paper provides a general purpose search-based technique for computing posterior probabilities in arbitrarily structured discrete[1] Bayesian networks. It is intended to be used for the case where there are extreme probabilities (see [Poole, 1993]). This paper shows how a major problem of practical efficiency can be solved by the use of a probabilistic analogue of the 'conflict' used in consistency-based diagnosis.

The main contributions of this paper are:

1. For the Bayesian net community, this paper provides a new search-based mechanism for computing probabilities in discrete Bayesian networks, that has practical significance for networks with extreme probabilities (i.e., each conditional probability is close to one or zero). This has been tested for networks with tens of thousands of nodes.

2. For the model-based diagnosis community, this paper provides a new representation for problems that is more general and more natural than previous representations. The algorithm gives a way to determine the accuracy of probability estimates.

3. It provides a way to bring the model-based diagnosis and probabilistic diagnosis communities together, with a common set of problems, and a common set of solutions.

Implementations of Bayesian networks have been placed into three classes [Pearl, 1988; Henrion, 1990]:

1. Exact methods that exploit the structure of the network to allow efficient propagation of evidence [Pearl, 1988; Lauritzen and Spiegelhalter, 1988; Jensen *et al.*, 1990].

2. Stochastic simulation methods that give estimates of probabilities by generating samples of instantiations of the network (e.g., [Henrion, 1988; Pearl, 1987]).

3. Search-based approximation techniques that search through a space of possible values to estimate probabilities (e.g., [Henrion, 1991; D'Ambrosio, 1992]).

---

[1] All of the variables have a finite set of possible values. We do not consider variables with an infinite set of possible values.

The method presented in this paper falls into the last class. While the efficient exact methods exploit aspects of the network structure, we instead exploit extreme probabilities to gain efficiency. The exact methods work well for sparse networks (e.g., are linear for singly-connected networks [Pearl, 1988]), but become inefficient when the networks become less sparse. They do not take the distributions into account. The method in this paper uses no information on the structure of the network, but rather has a niche for classes of problems where there are "normality"[2] conditions that dominate the probability tables (see Section 3). The algorithm is efficient for these classes of problems, but becomes very inefficient as the distributions become less extreme (see [Poole, 1993] for a detailed average-case complexity analysis of the simple version of the algorithm presented here (without conflicts)). This algorithm should thus be seen as having an orthogonal niche to the algorithms that exploit the structure for efficiency.

## 2 Background

A *Bayesian network* [Pearl, 1988] is a graphical representation of (in)dependence amongst random variables. A Bayesian network is a directed acyclic graph where the nodes represent random variables. If there is an arc from variable $B$ to variable $A$, $B$ is said to be a parent of $A$. The independence assumption of a Bayesian network says that each variable is independent of its non-descendents given its parents.

Suppose we have a Bayesian network with random variables $X_1, ..., X_n$. The parents of $X_i$ are written as $\Pi_{X_i} = \left\langle X_{i_1}, \cdots, X_{i_{k_i}} \right\rangle$. Suppose $vals(X_i)$ is the set of possible values of random variable $X_i$.

Associated with the Bayesian network are conditional probability tables which gives the conditional probabilities of the values of $X_i$ depending on the values of its parents $\Pi_{X_i}$. This consists of, for each $v_i \in vals(X_j)$ and $v_{i_j} \in vals(X_{i_j})$, probabilities of the form

$$P(X_i = v_i | X_{i_1} = v_{i_1} \wedge \cdots \wedge X_{i_{k_i}} = v_{i_{k_i}})$$

For any probability distribution, we can compute a joint distribution by

$$P(X_1, \cdots, X_n) = \prod_{i=1}^{n} P(X_i | \Pi_{X_i}).$$

This is often given as the formal definition of a Bayesian network.

We call an assignment of values to all the variables a *possible world*, and write '$\omega \models X_i = v_i$' if $X_i$ is assigned value $v_i$ in world $\omega$. Let $\Omega$ be the set of all possible worlds. The truth value of a formula (made up of assignments of values to variables and the standard logical connectives) in a possible world is determined using the standard truth tables.

## 3 Searching possible worlds

For a finite number of variables with a finite number of values, we can compute the probabilities directly, by enumerating the possible worlds. This is however computationally expensive as there are exponentially many of these (the product of the sizes of the domains of the variables).

The idea behind the search method presented in this paper can be obtained by considering the questions:

- Can we estimate the probabilities by only enumerating a few of the possible worlds?
- How can we enumerate just a few of the most probable possible worlds?
- Can we estimate the error in our probabilities?
- How can we make this search efficient?

This paper sets out to answer these questions, for the case where the distribution is given in terms of Bayesian networks.

### 3.1 Ordering the variables

The first thing we do is to impose a total ordering on the variables that is consistent with the ordering of the Bayesian network. We index the random variables $X_1, ..., X_n$ so that the parents of a node have a lower index than the node. This can always be done as the nodes in a Bayesian network form a partial ordering. If the parents of $X_i$ are $\Pi_{X_i} = \left\langle X_{i_1}, \cdots, X_{i_{k_i}} \right\rangle$, the total ordering preserves $i_j < i$.

The reason that we are interested in this ordering is that we can determine the probability of any formula given just the predecessors of the variables in the total ordering (as the parents of variables are amongst their predecessors).

### 3.2 Search Tree

We are now in a position to determine a search tree for Bayesian networks[3].

**Definition 3.1** A *partial description* is a tuple of values $\langle v_1, \cdots, v_j \rangle$ where each $v_i$ is an element of the domain of variable $X_i$.

The search tree has nodes labelled with partial descriptions, and is defined as follows:

- The root of the tree is labelled with the empty tuple $\langle \rangle$ (where $j = 0$).

---

[2]This should not be confused with "normal" as used for Gaussian distributions. We consider systems that have normal operating conditions and only rarely deviate from this normality (i.e., we are assuming abnormality [McCarthy, 1986] is rare). As we are only considering discrete variables, there should be no confusion.

[3]This search tree is the same as the probability tree of [Howard and Matheson, 1981] and corresponds to the semantic trees used in theorem proving [Chang and Lee, 1973, Section 4.4], but with random variables instead of complementary literals.

```
Q := {⟨⟩};
W := {};
While Q ≠ {} do
    choose and remove ⟨v₁,···,vⱼ⟩ from Q;
    if j = n
      then W := W ∪ {⟨v₁,···,vⱼ⟩}
      else Q := Q ∪ {⟨v₁,···,vⱼ,v⟩ : v ∈ vals(Xⱼ₊₁)}
```

Figure 1: Basic search algorithm

- The children of node labelled with $\langle v_1,\cdots,v_j \rangle$ are the nodes labelled with $\langle v_1,\cdots,v_j,v \rangle$ for each $v \in vals(X_{j+1})$. In other words, the children of a node correspond to the possible values of the next variable in the total ordering.

- The leaves of the tree are tuples of the form $\langle v_1,\cdots,v_n \rangle$. These correspond to possible worlds.

Tuple $\langle v_1,\cdots,v_j \rangle$ corresponds to the variable assignment $X_1 = v_1 \wedge \cdots \wedge X_j = v_j$.

We associate a probability with each node in the tree. The probability of the node labelled with $\langle v_1,\cdots,v_j \rangle$ is the probability of the corresponding proposition which is

$$P(X_1 = v_1 \wedge \cdots \wedge X_j = v_j)$$
$$= \prod_{i=1}^{j} P(X_i = v_i | X_{i_1} = v_{i_1} \wedge \cdots \wedge X_{i_{k_i}} = v_{i_{k_i}})$$

This is easy to compute as, by the ordering of the variables, all of the ancestors of every node have a value in the partial description.

The following lemma can be trivially proved, and is the basis for the search algorithm.

**Lemma 3.2** *The probability of a node is equal to the sum of the probabilities of the leaves that are descendents of the node.*

This lemma lets us bound the probabilities of possible worlds by only generating a few of the possible worlds and placing bounds on the sizes of the possible worlds we have not generated.

### 3.3 Searching the Search Tree

To implement the computation of probabilities, we carry out a search on the search tree, and generate some of the most likely possible worlds. Figure 1 gives a generic search algorithm that can be varied by changing which element is chosen from the queue. There are many different search methods that can be used [Pearl, 1984].

The idea of the algorithm is that there is a priority queue $Q$ of nodes. We remove one node at any time, either it is a total description (i.e., where $j = n$) in which case it is added to $W$, the set of generated worlds, or else its children are added to the queue.

Note that each partial description can only be generated once. There is no need to check for multiple paths or loops in the search. This simplifies the search, in that we do not need to keep track of a CLOSED list or check whether nodes are already on the OPEN list ($Q$ in Figure 1) [Pearl, 1984].

No matter which element is chosen from the queue at each time, this algorithm halts and when it halts $W$ is the set of all tuples corresponding to possible worlds.

## 4 Estimating the Probabilities

If we let the above algorithm run to completion we have an exponential algorithm for enumerating the possible worlds that can be used for computing the prior probability of any proposition or conjunction of propositions. This is not, however, the point of this algorithm. The idea is that we want to stop the algorithm part way through, and use the set of possible worlds generated to estimate the probabilities we need.

We use $W$, at the start of an iteration of the while loop, as an approximation to the set of all possible worlds. This can be done irrespective of the search strategy used.

### 4.1 Prior Probabilities

Suppose we want to compute $P(g)$. At any stage (at the start of the while loop), the possible worlds can be divided into those that are in $W$ and those that will be generated from $Q$.

$$P(g) = \sum_{w \in \Omega \wedge w \models g} P(w)$$
$$= \left( \sum_{w \in W \wedge w \models g} P(w) \right)$$
$$+ \left( \sum_{w \text{ to be generated from } Q : w \models g} P(w) \right)$$

We can easily compute the first of these sums, and can bound the second. The second sum is greater than zero and is less than the sum of the probabilities of the partial descriptions on the queue (using Lemma 3.2). This means that we can bound the probabilities of a proposition based on enumerating just some of the possible worlds. Let

$$P_W^g = \sum_{w \in W \wedge w \models g} P(w)$$

$$P_Q = \sum_{t \in Q} P(t)$$

**Lemma 4.1**

$$P_W^g \leq P(g) \leq P_W^g + P_Q$$

As the computation progresses, the probability mass in the queue $P_Q$ approaches zero and we get a better refinements on the value of $P(g)$. Note that $P_Q$ is monotonically non-increasing through the loop (i.e $P_Q$ stays the same or gets smaller through the loop). This thus forms the basis of an "anytime" algorithm for Bayesian networks.

### 4.2 Posterior Probabilities

The above analysis was for finding the prior probability of any proposition. If we want to compute the posterior probability of some $g$ given some observations $obs$, we can use the definition of conditional probability, and use

$$P(g|obs) = \frac{P(g \wedge obs)}{P(obs)}$$

Given estimates of $P(g \wedge obs)$ and $P(obs)$, (namely $P_W^{g \wedge obs}$ and $P_W^{obs}$), it can be proved [Poole, 1993] that $P(g|obs)$ has the following bound:

**Theorem 4.2**

$$\frac{P_W^{g \wedge obs}}{P_W^{obs} + P_Q} \leq P(g|obs) \leq \frac{P_W^{g \wedge obs} + P_Q}{P_W^{obs} + P_Q}$$

If we choose the midpoint as an estimate, the maximum error is

$$\frac{1}{2} \left( \frac{P_W^{g \wedge obs} + P_Q}{P_W^{obs} + P_Q} - \frac{P_W^{g \wedge obs}}{P_W^{obs} + P_Q} \right)$$
$$= \frac{P_Q}{2(P_W^{obs} + P_Q)}$$

What is interesting about this is that the error is independent of $g$. Thus when we are generating possible worlds for some observation, and want to have posterior estimates within some error, we can generate the required possible worlds independently of the proposition that we want to compute the probability of.

## 5 Discussion

### 5.1 Refinements to the Search Algorithm

There are a number of refinements that can be carried out to the algorithm of Figure 1, independently of the search strategy. Some of these are straightforward, and work well. The most straightforward refinements are:

- If we know our query and the conditioning variables, we can prune those variables that cannot affect the answers. We can prune any variables that are d-separated from the query variables by the observations [Pearl, 1988]. We can prune any variable that is not an ancestor of the observations or the query. The above two pruning steps can be done repeatedly [Baker and Boult, 1990]

- If we are trying to determine the value of $P(\alpha)$, we can stop enumerating the partial descriptions once it can be determined whether $\alpha$ is true in that partial description. When conditioning on our observations we can prune any partial description that is inconsistent with the observations.

- Another alternative is an iterative-deepening search [Korf, 1985]. As we are not concerned with finding the most likely possible world, but a set of most likely worlds, we can carry out depth-bounded depth-first searches (not generating nodes whose probability is below a threshold), without worrying too much about decreasing the threshold to the maximum value it could obtain.

### 5.2 Extreme Probabilities

The improvements to the search algorithm below assume we have extreme probabilities. This means that for each variable $X_i$ and for each instantiation of the parents of $X_i$, there is one value $v_i$ for which

$$P(X_i = v_i | X_{i_1} = v_{i_1} \wedge \cdots \wedge X_{i_{k_i}} = v_{i_{k_i}})$$

has a probability close to one (and so the conditional probabilities of the other values for $X_i$ are all close to zero).

**Definition 5.1** Variable $X_i$ is **normal** in possible world $\omega$ if $P(X_i = v_i | X_{i_1} = v_{i_1} \wedge \cdots \wedge X_{i_{k_i}} = v_{i_{k_i}}) \approx 1$ where $\omega \models X_i = v_i$ and $\omega \models X_{i_j} = v_{i_j}$ for $0 < j \leq k_i$. Otherwise we say $X_i$ is a **fault** in possible world $\omega$.

See [Poole, 1993] for a more detailed discussion of the use of extreme probabilities, and why these extreme probabilities guarantee convergence of the search.

## 6 A Diagnosis Example

In this section we describe how the search procedure can be applied to a simple circuit diagnosis problem (as in [de Kleer, 1991]), from which we can learn what problems arise. The translation of the circuit into a Bayesian network will follow that of Pearl [1988, Section 5.4].

The circuit is a sequence of a one-bit adders, cascaded to form a multiple-bit adder[4].

### 6.1 Representation

Figure 2 shows a one bit adder. Figure 3 shows the corresponding Bayesian network.

In this Bayesian network the random variable $out$-$a2$ is a binary random variable that has two values $on$ meaning that the output of gate $a2$ is on, and *off* meaning the output of the gate $a2$ is off. The random variable $a2ok$ has three values: $ok$ meaning that the gate $a2$ is working correctly, $stuck1$ meaning the gate $a2$ is broken, and always has *on* and $stuck0$ meaning the gate $a2$ is broken, and always has *off*.

---

[4]There is actually an efficient algorithm for such an example using a clique hypertree representation [Lauritzen and Spiegelhalter, 1988; Jensen et al., 1990]. This exploits the local nature of the propagation, which we do not exploit. These would not work so well when the structure cannot be exploited as well as for the cascaded adders, for example, if we add to the circuit another circuit to find the parity of the resulting bits. We chose this example as it is simple to extend to large systems and also because it was used in [de Kleer, 1991].

| $a2ok$ | $i3$ | $out\text{-}x1$ | $out\text{-}a2$ | |
|---|---|---|---|---|
| | | | on | off |
| $ok$ | on | on | 1 | 0 |
| $ok$ | on | off | 0 | 1 |
| $ok$ | off | on | 0 | 1 |
| $ok$ | off | off | 0 | 1 |
| $stuck1$ | – | – | 1 | 0 |
| $stuck0$ | – | – | 0 | 1 |

Figure 4: Conditional probability table for variable $out\text{-}a2$.

| $a2ok$ | | |
|---|---|---|
| $ok$ | $stuck1$ | $stuck0$ |
| 0.99999 | 0.000005 | 0.000005 |

Figure 5: Conditional probability table for variable $a2ok$.

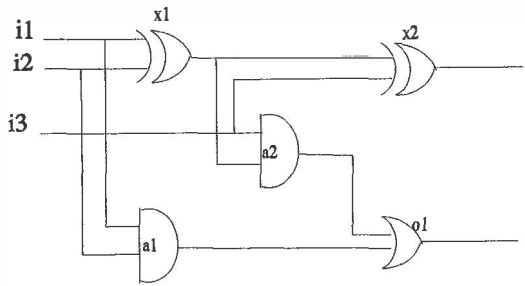

Figure 2: 1 bit adder

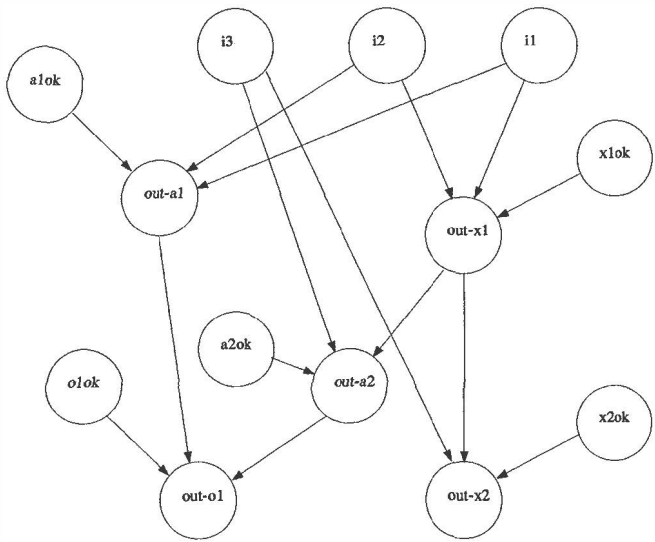

Figure 3: Bayesian network for a 1 bit adder

The value of $out\text{-}a2$ depends on the values of three other variables, $i3$, $out\text{-}x1$, and $a2ok$. The values for the variable $out\text{-}a2$ follow the table in Figure 4. The tables for the other outputs of gates is similar.

The value of $a2ok$ does not depend on any other variables. The values for the variable follow the table in Figure 5.[5] The tables for the status of other gates is similar.

These one-bit adders can be cascaded for form multiple bit adders. This is done in the circuit by connecting the output of gate $o1$ in one adder to input $i3$ of the following adder. In the Bayesian network, this is done by having multiple instances of the network for the one-bit adder with the value of $i3$ depending on the variable $out\text{-}o1$ for the previous instance of the adder. The table for the probabilities is given in figure 6. The value of the output of gate $x2$ of bit $k$, is called the *output* of bit $k$; the value of the output of $o1$ is called the *carry*.

### 6.2 Computation

Suppose we apply the algorithm of Figure 1 to our cascaded adder example with the partial description with the highest prior probability chosen each time through the loop. First the world with all gates being $ok$ is generated followed by the worlds with single faults. Most of these can be pruned quickly (see Section 5.1). Then the double stuck-at faults are generated, etc. The probability in the queue converges very quickly [Poole, 1993]. Each of the elements of the queue can be characterized by what errors are in the partial description. We typically only generate the partial descriptions with only a few of the errors.

---

[5]The numbers are purely made up. It may seem as though these probabilities are very extreme, but a 1000 bit adder (with 5000 components), is only 95% reliable, if all of the gates are as reliable as that given in this table.

| $out\text{-}o1_{k-1}$ | $i3_k$ | |
|---|---|---|
| | on | off |
| on | 1 | 0 |
| off | 0 | 1 |

Figure 6: Conditional probability table for input 3 of adder $k$.

This is essentially the candidate generator phase of [de Kleer, 1991]. From this candidate generation, we can compute all of the probabilities that we need to.

To see what computational problem arises, consider a 1000-bit adder. Suppose all the inputs are zero, and all outputs, except bit $k$, are zero, and bit $k$ outputs one (this example is from [de Kleer, 1991]). For $k > 1$ there are five most likely possible worlds (that correspond to $x2ok_k = stuck1$, $x1ok_k = stuck1$, $o1ok_{k-1} = stuck1$, $a2ok_{k-1} = stuck1$, and $a1ok_{k-1} = stuck1$). We first choose the most likely values of all variables (i.e., the $ok$ state for all of the status nodes). When we get to the output of bit $k$, which is predicted to be zero, we find that our prediction is inconsistent with the observations. At this stage, we prune the search and consider the single-fault possible worlds. For each bit after bit $k$, we have already assigned a single fault (to account for the error in bit $k$), thus for each of these gates, we only consider the $ok$ state. For all of the gates before bit $k$, we consider each of the failure states. When generating worlds with just single faults, there is no point in trying each of the failure states for the gates before bit $k - 1$ as each of these failure states will have to be combined with another failure state to account for the error. We would like to not consider faults that we know will have to be combined with other faults until we need to (when considering double fault worlds these may have to be considered). Learning what we can about expectation failure and using this information for pruning the search is the idea behind the use of conflicts.

## 7 Search Strategy and Conflicts

The above example assumed a simple search strategy. We can carry out various search strategies, to enumerate the most likely possible worlds. Here we present one that incorporates an analogous notion of conflict to that used in consistency-based diagnostic community [de Kleer and Williams, 1987; Reiter, 1987; de Kleer et al., 1990].

We carry out a multiplicative version[6] of $A^*$ search [Pearl, 1984] by choosing the node $m$ from the queue with the highest value of $f(m) = g(m) \times h(m)$. Here $g(m)$ is the

---

[6]This is an instance of $Z^*$ where, instead of adding the costs and choosing the minimum we multiply and choose the maximum. This can be transformed into a more traditional $A^*$ algorithm by taking the negative of the logarithms of the probabilities. We do not do this explicitly as we want the probabilities to add after the search.

probability of the corresponding proposition:

$$\begin{aligned}
g(\langle v_1, \cdots, v_j \rangle) \\
&= P(X_1 = v_1 \wedge \cdots \wedge X_j = v_j) \\
&= \prod_{i=1}^{j} P(X_i = v_i | X_{i_1} = v_{i_1} \wedge \cdots \wedge X_{i_{k_i}} = v_{i_{k_i}}) \\
&= P(X_j = v_j | X_{j_1} = v_{j_1} \wedge \cdots \wedge X_{j_{k_j}} = v_{j_{k_j}}) \\
&\quad \times g(\langle v_1, \cdots, v_{j-1} \rangle)
\end{aligned}$$

The heuristic function $h(\langle v_1, \cdots, v_j \rangle)$ is the product of the maximum probabilities that can be obtained by variables $X_{j+1} \cdots X_n$ (for any values of the predecessors of these variables). Initially, these can be computed by a linear scan (from $X_n$ to $X_1$) keeping a table of the maximum products. We use a notion of conflicts to refine the heuristic function as computation progresses. This is defined in terms of normality (Section 5.2), and is closest to that of [Reiter, 1987].

**Definition 7.1** Given an observation $o$, a **conflict** is a set $C$ of random variables such that there is no possible world in which $o$ is true and all elements of $C$ are normal. In other words, if $o$ is true, one of the elements of $C$ is a fault (and so has probability close to zero, no matter what values are assigned to variable outside of $C$).

Associated with a conflict is a **maximum probability** which is an upper bound on the prior probability of any assignment of values to variables in the conflict that is consistent with the observation.

Two conflicts are **independent** if there is no single variable that can account for both conflicts. That is, $C_1$ and $C_2$ are independent if in all possible worlds in which the observation is true there are at least two faults, one in $C_1$ and one in $C_2$. This happens, for example, if the conflicts have no variables in common.

**Example 7.2** In our example of Section 6, with all inputs zero, and bit 50 having output one and all other outputs being zero, there is one minimal conflict, namely:
$\{out\text{-}x2_{50}, x2ok_{50}, i3_{50}, out\text{-}o1_{49}, o1ok_{49}, out\text{-}a1_{49},$
$a1ok_{49}, i2_{49}, out\text{-}a2_{49}, a2ok_{49}, out\text{-}x1_{49}, x1ok_{49}, i1_{49}, i2_{49},$
$out\text{-}x1_{50}, x1ok_{50}, i1_{50}, i2_{50}\}$.

A conflict corresponds to a set of normal values that cannot consistently coincide given the observation. Conflicts discovered in the search can be used to prune the search earlier than it could be pruned without the conflict. There are a number of issues to be discussed:

1. How can conflicts be used by the search algorithm?
2. How can conflicts be discovered?
3. How does the use of conflicts affect the estimation of probabilities?
4. How much does the use of conflicts save in search time?
5. In practice, how often can we detect a small set of variables that form a conflict?

In this paper we answer all but the last of these questions. The last question we cannot answer until we have built many more systems for many diverse applications.

### 7.1 Refining the heuristic function

We can use a conflict to update the heuristic function. In particular a conflict can change the bound on the probability that the rest of the variables can take on.

The simplest idea is that $h(\langle v_1, \cdots, v_j \rangle)$ is the product of the maximum probabilities of the independent conflicts that involve only variables after variable $X_j$[7].

A discovered conflict updates the heuristic function for all the variables before (in the total variable ordering) the conflict. The heuristic function evolves as computation progresses and conflicts are found.

### 7.2 Finding conflicts

We would like a way to extract conflicts from expectation failures in our search. By the nature of the search, we first consider the most likely values of variables. We only want to consider faults that are forced upon us by conflicts.

Suppose that in the current partial description variable $X_i$ is assigned a value $v_i$ and it has been observed that $X_i = o$ where $o \neq v_i$. We say that the value $v_i$ is predicted, and we have a failure of expectation. We would like to extract a conflict from the current partial description.

We want to extract conflicts as fast as possible, and do not necessarily want to build the whole infrastructure of our diagnosis system around finding conflicts (as does de Kleer [1991]). We would like to extract the conflicts from the expectation failure directly. We are not particularly concerned about finding minimal conflicts[8]. Whether these are reasonable design goals, or achievable in practice remains to be seen.

A set of variables $C$ is a **counter** to $X_i = o$ if there is no possible world in which every variable in $C$ is normal and $X_i = o$. A counter to a conjunction of assignments to variables is defined analogously.

To generate a counter to $X_i = o$, we consider each tuple $\langle v_{i_1}, \cdots, v_{i_{k_i}} \rangle$ of values to the parents such that $P(X_i = o | X_{i_1} = v_{i_1} \wedge \cdots \wedge X_{i_{k_i}} = v_{i_{k_i}}) \approx 1$. A counter to $X_i = o$ must contain a counter to each of these conjunctions of assignments of values to the parents of $X_i$.

The idea of the algorithm $extract\_counter$ that finds counters is that we recursively find counters of these assignments to the parents, union them, add $X_i$ and return this set as a counter to $X_i = o$.

The problem is how to find the counter to the conjunction, without doing lots of search. This is where the extraction from the failure of an expectation comes into play. For each conjunction, there was a conjunct whose negation was predicted in the current partial description (otherwise $X_i$ would have been predicted to have value $o$). We use the procedure recursively to extract a counter to that assignment from the current partial description.

The procedure $extract\_counter(X_i, o, \delta)$ where $X_i$ is a variable, $o$ is a value and $\delta$ is a partial description such that $X_i = o$ is not true in $\delta$, is defined as follows. Suppose the parents of $X_i$ are $\langle X_{i_1}, \cdots, X_{i_{k_i}} \rangle$. Consider each tuple $\langle v_{i_1}, \cdots, v_{i_{k_i}} \rangle$ of values to the parents such that $P(X_i = o | X_{i_1} = v_{i_1} \wedge \cdots \wedge X_{i_{k_i}} = v_{i_{k_i}}) \approx 1$. Choose[9] $i_j$ such that $X_{i_j} \neq v_{i_j}$ in the current partial description (i.e. $X_{i_j} = v_{i_j}$ is not predicted). Recursively call $extract\_counter(X_{i_j}, v_{i_j}, \delta)$. This returns the set of variables all of which cannot be normal if $X_{i_j} = v_{i_j}$. The value returned for $extract\_counter(X_i, o, \delta)$ is then

$$\{X_i\} \cup \bigcup_{\substack{P(X_i = o | X_{i_1} = v_{i_1} \wedge \cdots \wedge X_{i_{k_i}} = v_{i_{k_i}}) \approx 1 \\ \text{and } \delta \models X_{i_j} \neq v_{i_j}}} extract\_counter(X_{i_j}, v_{i_j}, \delta)$$

So when we have a failure of expectation caused by the observation $X_i = o$, then $extract\_counter(X_i, o, \delta)$ will return a conflict.

N.B. sometimes $extract\_counter$ may fail to find a counter if $\delta$ contains a fault that produces the expectation failure. In this case we cannot extract a conflict that is independent of the conflict that forced the fault in $\delta$.

### 7.3 Estimating probabilities

Naive use of the above procedure gives error estimates that are too large. The problem is that there are quite large probabilities on the queue that are not chosen because of their heuristic values. Thus the value of $P_Q$ is much larger than we may expect.

Suppose $m$ is an element on the queue, that is not chosen because $f(m) = g(m) \times h(m)$ is too low. Although the set of possible world rooted at $m$ has probability $g(m)$, most of these are impossible if there is a conflict. We know at least $(1 - h(m))$ of the weighted sum of these possible worlds must be inconsistent with the observations (by the definition of $h$). This we should only count $f(m) = g(m) \times h(m)$ as part of $P_Q$, rather than $g(m)$. This can then be used to estimate probabilities, and gives a much better accuracy.

### 7.4 Experimental Results

The experiments we carried out were limited to understanding the behaviour of the algorithm on cascaded $n$-bit adder example, with all inputs $zero$ and all output bits being $zero$, except for the output of bit $k$ (i.e., the value of $x2_k$) which had value $one$. Note that an $n$-bit adder has $5n$ gates and corresponds to a Bayesian network with $13n$ nodes. We ran the program using a bounded depth-first search (pruning the depth-first search when the $f$-value gets below a threshold), generating the 5 most likely possible worlds.

---

[7] A more sophisticated version may count conflicts that contain variables before $X_j$, (and do not include $X_j$) as long as they are assigned normal values in $\langle v_1, \cdots, v_j \rangle$. We have only tested the simpler idea.

[8] Correctness does not depend on a conflict being minimal.

[9] Any choice will lead to a conflict. A bad choice may lead to a non-minimal conflict. Our experiments were with a greedy algorithm that chooses the first one found. There is a tradeoff between the computational effort in finding minimal conflicts, and the extra pruning that minimal conflicts allow.

| error bit | 2 | 25 | 50 | 75 | 100 |
|---|---|---|---|---|---|
| run time (no conflicts) | 14 | 56 | 188 | 408 | 718 |
| run time (with conflicts) | 16 | 13 | 10 | 7 | 4 |

Figure 7: Running time as a function of error bit in a 100-bit adder.

| # bits | 100 | 500 | 1000 | 2000 | 3000 |
|---|---|---|---|---|---|
| # gates | 500 | 2500 | 5000 | 10000 | 15000 |
| # nodes | 1300 | 6500 | 13000 | 26000 | 39000 |
| run time | 10 | 46 | 92 | 183 | 275 |

Figure 8: Running time as a function of size of multiple-bit adder.

All times are based on a SICStus Prolog program running on a NeXTstation. All times are in seconds. The code is available from the author.

The main problem with the search algorithm without conflicts, for our example, was how the runtime depended on the bit $k$ that was faulty. Figure 7 shows how run time depends on the bit chosen for the program with no conflicts and for the program with conflicts. This was for the 100-bit adder (Bayesian network with 1300 nodes). The difference in times for error bit 2 indicates the overhead in using conflicts (as conflicts for this case gives us nothing).

Consider how the program runs: we pursue one world until bit $k$, then pursue 5 worlds separately from bits $k$ to $n$. Thus we may estimate the time as proportional to $k + 5(n - k)$. This fits the experimental results extremely well.

The second experiment was with the asymptotic behaviour as the size of the network was increased. Figure 8 shows the run-time for finding the 5 most likely possible worlds, as a function of circuit size. In each of these the error bit was the middle bit of the circuit (i.e., $k = \frac{n}{2}$). This was chosen as it is the average time over all of the error bits (see Figure 7). Note the linear time that was predicted by the $k + 5(n - k)$ formula.

Finally, the results from double errors, are very similar. For a 100-bit adder, with ones observed at bits 30 and 70, the program took 34 seconds to find the 25 most likely possible worlds.

## 8 Comparison with other systems

The branch and bound search is very similar to the candidate enumeration of de Kleer's focusing mechanism [de Kleer, 1991]. We have considered a purely probabilistic version of de Kleer's conflicts. We have extended the language to be for Bayesian networks, rather than for the more restricted and less well-defined language that de Kleer uses. We also can bound the errors in our probabilistic estimates, which de Kleer cannot do. One of the features of our work is that finding minimal conflicts is not essential to the correctness of the program, but only to the efficiency. Thus we can explore the idea of saving time by finding useful, but non-minimal conflicts quickly.

Shimony and Charniak [1990], Poole [1992a] and D'Ambrosio [1992] have proposed back-chaining search algorithms for Bayesian networks. None of these are nearly as efficient as the one presented here. Even if we consider finding the single most normal world, the algorithm here corresponds to forward chaining on definite clauses (see [Poole, 1992b]), which can be done in linear time, but backward chaining has to search and takes potentially exponential time.

This paper should be seen as a dual to the TOP-N algorithm of Henrion [1991]. We have a different niche. We take no account of the noisy-OR distribution that Henrion concentrates on.

This paper deliberately takes the extreme position of seeing how far we can get when we exploit the distributions and not the structure of the network. Hopefully this can shed light on the algorithms that use both the structure and the distribution to gain efficiency (e.g., [D'Ambrosio, 1992]).

## 9 Conclusion

This paper has considered a simple search strategy for computing prior and posterior probabilities in Bayesian networks. This uses a variation on $A^*$ search, and uses a notion of 'conflict' to refine the heuristic function. One of the aims of this work is to bring together the model-based diagnosis community (e.g., [de Kleer, 1991]) and the uncertainty in AI community, with a common set of problems and tools.

In some sense this is preliminary work. We have not tested this beyond the single example. It is not clear how easy it will be in other examples to find conflicts without searching for counters, nor how much the use of conflicts can save us. The use of counters seems to be very different to exploitation of structure in other algorithms, but there may be some, as yet undiscovered, relationship there.

## Acknowledgements

Thanks to Craig Boutilier, Nevin Zhang, Runping Qi and Michael Horsch for valuable comments on this paper. This research was supported under NSERC grant OGPOO44121, and under Project B5 of the Institute for Robotics and Intelligent Systems.